\documentclass[10pt,twocolumn,letterpaper]{article}
\usepackage{cvpr}
\usepackage{times}
\usepackage{epsfig}
\usepackage{graphicx}
\usepackage{amsmath}
\usepackage{array}
\usepackage{kantlipsum}
\usepackage{amssymb}
\usepackage{float}
\usepackage{mathpazo}
\usepackage{authblk}
\usepackage{tabularx} 
\usepackage{algorithm2e}
\usepackage{enumitem}
\usepackage{tabulary}
\usepackage{adjustbox}
\usepackage{lipsum}

 \cvprfinalcopy 
\def\cvprPaperID{8214} 

\usepackage[utf8]{inputenc}
\DeclareUnicodeCharacter{2212}{-}
\begin{document}
\pagenumbering{arabic}

\title{Self-Attention Capsule Networks for Object Classification}
\author{Assaf Hoogi$^{1,2}$, Brian Wilcox $^{3*}$, Yachee Gupta$^{1*}$, Daniel L. Rubin$^{1}$ \\ \small{$^{1}$ Dept. of Biomedical Data Science, Stanford University \\$^{2}$ Dept. of Computer Science and Applied Math, The Weizmann Institute of Science \\$^{3}$ Dept. of Electrical Engineering, Stanford University \\$^{*}$ Equal contributors}}

\maketitle

\begin{abstract}
We propose a novel architecture for object classification, called Self-Attention Capsule Networks (SACN). SACN is the first model that incorporates the Self-Attention mechanism as an integral layer within the Capsule Network (CapsNet). While the Self-Attention mechanism supplies a long-range dependencies, results in selecting the more dominant image regions to focus on, the CapsNet analyzes the relevant features and their spatial correlations inside these regions only. The features are extracted in the convolutional layer. Then, the Self-Attention layer learns to suppress irrelevant regions based on features analysis and highlights salient features useful for a specific task. The attention map is then fed into the CapsNet primary layer that is followed by a classification layer. The proposed SACN model was designed to solve two main limitations of the baseline CapsNet - analysis of complex data and significant computational load. In this work, we use a shallow CapsNet architecture and compensates for the absence of a deeper network by using the Self-Attention module to significantly improve the results. The proposed Self-Attention CapsNet architecture was extensively evaluated on six different datasets, mainly on three different medical sets, in addition to the natural MNIST, SVHN and CIFAR10. The model was able to classify images and their patches with diverse and complex backgrounds better than the baseline CapsNet. As a result, the proposed Self-Attention CapsNet significantly improved classification performance within and across different datasets and outperformed the baseline CapsNet, ResNet-18 and DenseNet-40 not only in classification accuracy but also in robustness.
\end{abstract}

\section{Introduction}
\begin{figure}[!t]
  \includegraphics[width=\columnwidth]{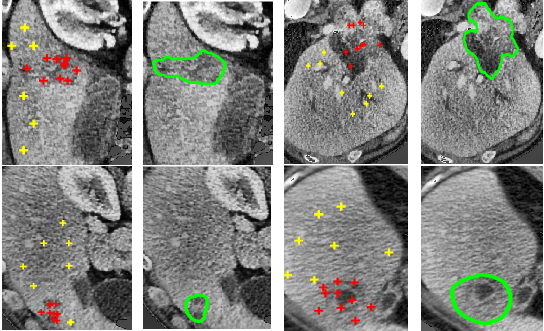}
      \label{LiTS2}
      \caption{Classification of \textbf {random patch examples} - CT Liver lesions (LiTS public). Each pair of images (in the same row) represent the original image with the radiologist's lesion annotation (green) and the processed image. Red - classified lesion patches, Yellow - classified normal patches. The dataset contains difficult cases such as low contrast and highly heterogeneous lesions.}
      \vspace{-2mm}
\end{figure}

Object classification is a very challenging task, mostly because of the significant intra-class and inter-class variability \ref{LiTS2}, arising from different image acquisition conditions, rigid and non-rigid deformations, occlusions and corruptions. Handcrafted low-level features were proposed to handle these challenges, while unsupervised learning approaches are regularly developed to avoid the limitations of handcrafted features such as being user dependant. 

Recent advances in computer vision highlight the capabilities of deep learning approaches to solve these challenges, achieving state of the art performances in many classification tasks. The main reason for the success of deep learning is the ability of Convolutional Neural Networks to learn a hierarchical representation of the input data. AlexNet, which was presented by Krizhevsky et al. \cite{NIPS2012_4824} was one of the first and the simplest architectures for object/image classification. Later on, a deeper VGG16 model was introduced, dealing with nonlinear transformations \cite{simonyan2014deep}. ResNet-18 was then developed to solve a common problem in deep learning of increased test error rate while increasing the architecture depth \cite{He_2016}. DenseNet-40 \cite{DenseNet} was recently developed and is currently considered as the state of the art method for object/image classification tasks. It is similar to the ResNet-18 architecture with the difference being significantly densely connected feature maps in the final layer of a dense block instead of a residual block.\newline

Deep learning-based approaches became popular also in medical image domain due to the increasing computational power and availability of data. However, these methods still lack in robustness across different datasets and require a significant amount of annotated data. These limitations are even more substantial in the medical domain (relative to the natural domain) because the annotated data is highly heterogeneous and its size is relatively small. To tackle these challenges, the U-Net architecture was developed. It includes skip connections and was designed for medical images, wherein these additional connections can extract larger amounts of information from the limited data size \cite{Ronneberger_2015} \cite{UN}. \newline
CapsNet is one of the most recent architectures that was developed by Hinton's group for object classification  \cite{sabour2017dynamic}. It is powerful and was designed to deal with small datasets, as is typical for the medical domain. It learns the spatial correlations between objects and can recognize multiple objects in the image even if they overlap. However, CapsNet does not learn the important local features. Therefore, developing a new architecture that will help to solve the mentioned limitations is highly desired and can help in advancing the field of object classification, especially in the medical domain. \newline

\noindent This paper presents \textbf{a significant improvement of the Capsule Networks architecture} and has several key contributions.
\begin{itemize}[label={\tiny\raisebox{1ex}{\textbullet}}]
      
   \item We introduce a novel architecture, called \textbf{Self-Attention Capsule Networks (SACN)}. The architecture includes an integral Self-Attention layer that lies between the convolutional and the primary CapsNet layers. This allows the model parameters, even in shallower layers, to be updated mostly based on image regions that are more relevant to a given task. The attention mechanism, which is used as a non-local operation, solves the task of learning the important features while the CapsNet considers the positional / rotational spatial relation between these features. Therefore, \textbf{integrating Self-Attention module within the CapsNet architecture has an important complementary role}.
   \item The proposed architecture is \textbf{designed to work well under the constraint of limited computational resources}. While the baseline CapsNet \cite{sabour2017dynamic} is considered an expensive architecture in terms of computational cost, the proposed model was designed to use a shallow CapsNet architecture and compensates the absence of a deeper network by using the Self-Attention module to significantly improve the feature extraction. \textbf{Our proposed SACN enjoys both worlds - it keeps computational efficiency while having a large receptive field at the same time, obtaining long-range dependencies} to improve performance. 
   \item We are not familiar with other works that were tested on both medical and natural image domains, as these domains have substantially different image characteristics. Here we conducted extensive experiments to show the \textbf{generalization of our model}. We showed that the proposed model was able to supply \textbf{more accurate, robust and stable object classification} within and across different datasets. Moreover, these datasets contains complex lesions such as low contrast lesions or lesions with high heterogeneity. \textbf{We were able to show that our proposed SACN model performs better than the baseline CapsNet on complex data, the weakest link of the baseline method}.   
\end{itemize}

\section {Related work}
\subsection {CapsNet architecture}
Recently developed Capsule networks represent a breakthrough in the field of neural networks. The CapsNet architecture contains three types of layers - the convolutional layer, the primary capsule layer and the classification (digit) capsule layer \cite{sabour2017dynamic}. Capsule networks are powerful because of two key ideas that distinguish them from the traditional CNNs; 1) dynamic routing-by-agreement instead of max pooling, and 2) squashing, where scalar output feature detectors of CNNs are replaced with vector output capsules.
Routing-by-agreement means that it is possible to selectively choose which parent in the layer above the  capsule is sent to. For each optional parent, the capsule network can increase or decrease the connection strength. As a result, the CapsNet can keep the spatial correlations between objects within the image \cite{sabour2017dynamic}.
Squashing means that instead of having individual neuron’s sent through non-linearities as is done in CNNs, capsule networks have their output squashed as an entire vector. The squashing function enables a better representation of the probability that an object is present in the input image. \textbf{This better representation, in addition to the fact that no max-pooling is applied (i.e. minimizing the information loss), enables CapsNet to handle small and sparse set of images, as is typical for medical imaging}. \newline
Afshar et al. \cite{Afshar_2018} incorporated CapsNets for brain tumor classification. The authors investigated the over-fitting problem of CapsNets based on a real set of MRI images. Their results show that CapsNet can successfully outperform CNNs for the brain tumor classification problem. 
However, CapsNet is an expensive architecture in terms of computational and memory loads. \textbf{As a result, the commonly-used CapsNets are relatively shallow architectures,  which were proved to be better mainly for simple datasets. They did not perform well for more complex data}. Deliege et al. \cite{delige2018hitnet} introduced HitNet, a deep learning network characterized by the use of a Hit-or-Miss layer composed of capsules. The idea is that the capsule corresponding to the true class has to make a hit in its target space, and the other capsules have to make misses. The method converged faster than CapsNet but their results were not able to outperform CapsNet for complex datasets. In \cite{Xi2017CapsuleNP}, the authors explored the effect of a variety of CapsNet modifications, ranging from stacking more capsule layers to trying out different parameters such as increasing the number of primary capsules or customizing an activation function. However, the best validation accuracy for a relatively complex dataset that their architecture reached was $71.55\%$, only comparable to CapsNet performance on the same dataset. They mentioned that computational resources limited their performance. Another architecture, Diverse Capsule Networks, introduced in \cite{phaye2018dense}, was able to supply only a $0.31\%$ improvement over the baseline CapsNet accuracy. 

\subsection {Self-Attention mechanism}
 The Self-Attention mechanism can help the model focus on more relevant regions inside the image and gain better performance for classification tasks with fewer data samples \cite{NIPS2014_5268} or more complex image backgrounds. Attention mechanism allows models to learn deeper correlations between objects \cite{mnih2014recurrent} and helps discover interesting new patterns within the data \cite{Jo2019QuantitativePI} \cite{olshausen:neurobiological}. Additionally, it helps in modeling long-range, multi-level dependencies across different image regions. Wang et al. \cite{wang2017nonlocal} address the specific problem of CNNs processing information too locally by introducing a Self-Attention mechanism, where the output of each activation is modulated by a subset of other activations. This helps the CNN to consider smaller parts of the image if necessary. Larochelle and Hinton \cite{NIPS2010_4089} proposed using Boltzmann Machines that choose where to look next to find locations of the most informative intra-class objects, even if they are far away in the image. Reichert et al. proposed a hierarchical model to show that certain aspects of attention can be modeled by Deep Boltzmann Machines \cite{Reichert2011AHG}. Attention-based models were also proposed for generative models. In \cite{tang2013learning}, the authors introduce a framework to infer the region of interest for generative models of faces. Their framework is able to pass the relevant information only, through the generative model. Recent technique that focuses on generative adversarial models is called SAGAN \cite{Goodfellow}. The authors proposed Self-Attention Generative Adversarial Networks (SAGAN) that achieve state-of the art generative results on the ImageNet dataset.\newline
 Recent work that deals specifically with medical data can be found in \cite{oktay2018attention}. This work presents an attention mechanism that is incorporated in the U-Net architecture for tissue/organ identification and localization. However, U-Net was mainly developed for segmentation tasks in the medical domain, rather than for object classification.
 
Our SACN model plays a key role in advancing the medical imaging, as most classification tasks in this domain need positional relationships between features to perform optimally. By using our architecture, we can focus the attention on relevant locations in the input and analyze the spatial relationships between their features by taking advantage of the CapsNet structure. 

 \begin{figure*}[t!]
  \includegraphics[width=\textwidth]{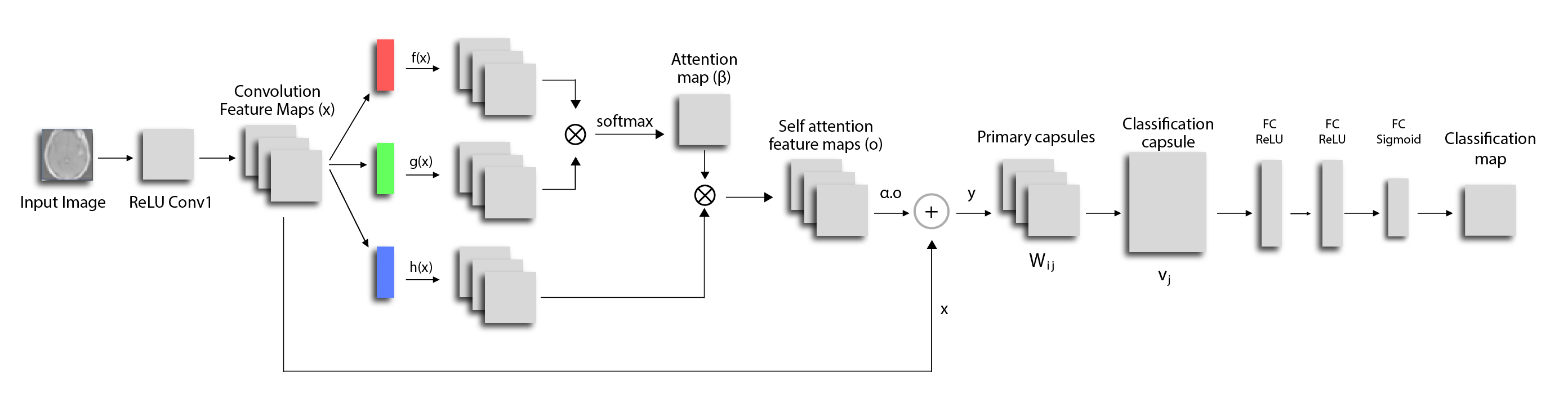}
  \caption{Our proposed SACN architecture.}
  \label{SelfAttentionCapsNet}
\end{figure*}

\section {The proposed model}
Our proposed model is illustrated in Figure \ref{SelfAttentionCapsNet}. Let $x \in R^{C\times N}$ be the output feature matrix extracted from the initial convolutional layer of the CapsNet, which is then fed into a Self-Attention module. Let $f(x),g(x)$ and $h(x)$ be three feature extractors. $h(x)$ has the same number of channels ($C$) as the input. We found that 512 channels supplies the best performance accuracy. $f(x)$ and $g(x)$ are position modules that are used to calculate  attention. $f(x_i)$ and $g(x_j)$ take input feature maps at the $i^{th}$ and $j^{th}$ positions. $f(x_i)$ and $g(x_j)$ both have a reduced number of channels ($C/8$), comparing with $h(x)$. This allows us to filter out noisy input channels and care only about the features that are relevant to the attention mechanism.  According to the dominant features, we pass only relevant activations through the Primary Capsule layer. Inside the attention module we use 2-D non-strided $1\times1$ convolutions and a non-local approach \cite{wang2017nonlocal}. This helps the CapsNet to model relationships between spatial regions that are far from each other and helps to keep the balance between efficiency and long-range dependencies (large receptive fields) by supplying a weighted sum of the features at all image locations. We define the non-local operation as:

\begin{equation}
 \eta_{ij}(x)=f(x_i)^T g(x_{j})
\end{equation}\newline
 $f(x_i)=W_fx_i, \hspace{5pt} g(x_i)=W_gx_i$, where $W_g \in R^{C \times C}$ and $W_f \in R^{C \times C}$ are the learned weight matrices. We then compute the softmax of $\eta_{ij}$ to get an output attention map $\beta_{ij}$,
 
 \begin{equation}
\beta_{ij}=\frac{exp(\eta_{ij})}{\sum_{i=1}^{N} exp(\eta_{ij})}
\end{equation}\newline
To obtain the final Self-Attention map, $o\in R^{C \times N}$, which will be the input of the primary CapsNet capsule, we apply matrix multiplication between the attention map $\beta_{ij}$ and $h(x_i)$,

 \begin{equation}
 o_j=\sum_{i=1}^{N}\beta_{ij} h(x_i)
\end{equation}\newline
where $h(x_i)=W_hx_i$ is the third input feature channel with $C$ channels (see Figure \ref{SelfAttentionCapsNet}) and similarly to $W_f$ and $W_g$, $W_h$ is also a learned weight matrix. By virtue of this matrix multiplication step, the Self-Attention mechanism applies a weighted sum over all derived features of $h(x_i)$ and filters out the ones that have the least affect. Therefore, the final output of the Self-Attention layer is 
 \begin{equation}
 y_i = \alpha o_i + x_i
\end{equation}
In our model, $\alpha$ is initialized to 0. As a result, the model can explore the local spatial information first, before automatically refining it with the Self-Attention and analyzing higher data complexity by considering further regions in the image. Then, the network gradually learns to assign higher weight to the non-local regions. By initializing $\alpha$ to $0$ and with no requirement of other pre-defined parameters, we are not dependent on the user input, contrary to common attention mechanisms.

The final output of the Self-Attention module, $y_i$, is then fed into the CapsNet primary layer. Let $v_j$ be the output vector of capsule $j$. The length of the vector, which represents the probability of whether or not a specific object is located in that given location in the image, should be between 0 and 1. To ensure that, we apply a squashing function that keeps the positional information of the object. Short vectors are shrunk to almost 0 length and long vectors are brought to a length slightly below 1.  The squashing function is defined as
\begin{equation}
v_j=\frac{||\sum_{i}c_{ij} W_{ij} y_{i}||^2}{(1+||\sum_{i}c_{ij} W_{ij} y_{i}||^2)}\frac{\sum_{i}c_{ij} W_{ij} y_{i}}{||\sum_{i}c_{ij} W_{ij} y_{i}||}
\end{equation}\newline
where $W_{ij}$ is a weight matrix and $c_{ij}$ are the coupling coefficients between capsule $i$ and all the capsules in the layer above $j$ that are determined by the iterative dynamic routing process 
\begin{equation}
c_{ij}=\frac{\exp{(b_{ij})}}{\sum_{j}\exp{(b_{ij})}}
\end{equation}\newline 
$b_{ij}$ are the log prior probabilities that $i^{th}$ capsule should be coupled to $j^{th}$ capsule.\newline 

To obtain a reconstructed image during training, we use the vector $v_j$ that supplies the highest coupling coefficient, $c_{ij}$. Then, we feed the correct $v_j$ through two fully connected ReLU layers. The reconstruction loss $L_R(I, \hat{I})$ of the architecture is defined as,

\begin{equation}
L_R(I, \hat{I})=|| I - \hat{I}||_{2}^{2}
\end{equation} 
where $I$ is the original input image and $\hat{I}$ is the reconstructed image. 
$L_R(I, \hat{I})$ is used as a regularizer that takes the output of the chosen $v_j$ and learns to reconstruct an image, with the loss function being the sum of squared differences between the outputs of the logistic units and the pixel intensities (L2-Norm). This forces capsules to learn features that are useful for the reconstruction procedure which inherently allows for the model to learn features at near-pixel precision. Therefore, the better the reconstruction loss the prediction. The reconstruction loss is then added to the following margin loss function, $L_M$, 

\begin{equation}
\begin{split}
L_M = &\sum_{k}T_k max(0,m^+ − ||v_k||)^2 +\\&\sum_{k}\lambda (1-T_k)max(0,||v_k||-m^-)^2
\end{split}
\end{equation}

$T_k=1$ if an instance of class k is present. $m^+=0.9$ and $m^-=0.1$ were selected as was suggested in \cite{sabour2017dynamic}.\newline
The end to end SACN architecture is evaluated and its weights are trained by using the total loss function, $L_T$, which is the total of all losses over all classes k,

\begin{equation}
L_T= L_M+\xi  I_{size} L_R
\end{equation}\newline
$\xi=0.0005$ is a regularization factor per channel pixel value that ensures that the reconstruction loss does not dominate over $L_M$ during training. $I_{size} = H*W*C$ is the number of input values, based on the height, width and number of channels in the input.

\subsection {Architecture and Hyper-parameters}
The input for the SACN architecture depends on the domain, the task and the data complexity. One of the main limitations of CapsNet refers to the analysis of complex data. Previous CapsNet research defined complex data as a data with a significant background heterogeneity.
For the medical domain we considered patch-wise CapsNet, similar to \cite{zbulak2019image}. Patch-wise analysis was chosen because of the desired task - local tissue classification. Moreover, by using patch-wise methodology we could also reduce the data complexity, to some extent. Patch size of $16\times16$ pixels was chosen as it supplied the best classification results. A value of $0.5$ was chosen for the $\lambda$ down-weighting of the loss, together with a weight variance of 0.15.
Since the CapsNet is a relatively expensive architecture, in terms of computational load, we designed our architecture to work well under the constraint of limited computational resources and boost the performance by adding the Self-Attention module. Therefore, our CapsNet architecture contains one convolutional layer with $5\times5$ filters, one Capsule layer and one routing iteration. 
In the Self-Attention module, all convolution layers use pixel-wise $1\times1$ kernels and spectral normalization to ensure that gradients are stable during training. We chose a batch size of 64 and a learning rate of $1e^{-3}$. Thirty epochs were used because this was sufficient to train the small dataset. \newline

Algorithm \ref{algorithm:SelfAttentionCapsNetAlgo} describes the training process of the proposed model. 

\begin{algorithm}
 \KwData{\textit{I},\textit{G}: Pairs of image \textit{I} and the ground truth \textit{G}}
 \KwResult{$Y_{out}$: Final instance classification}
 \While {not converging}{
 
  \textbf{CapsNet Convolutional Layer}: features are extracted and are divided into three output feature vectors ($f(x)$, $g(x)$, $h(x)$). 
  
   \textbf{Attention Layer}: Self-Attention map $y_i$ is generated based on the features vectors, attention map $\beta_{ij}$, learned weight matrix $W_h$ and a specific image location $x_i$.
 
    \textbf{Primary and Classification Layers}: the dominant features are then fed into the Primary CapsNet layer and from there to the Classification layer. Output classification $Y_{out}$ is obtained.
    
    \textbf{Calculate the Attention-based CapsNet loss}: $L_T \leftarrow Loss(G,Y_{out})$
    
     \textbf{Back-propagate the loss and compute}:  $\frac{\partial L}{\partial W}$
     
     \textbf{Update the weights}: matrices $W$ are updated for both the Self-Attention layer and the CapsNet architecture.
 }
 
 \caption{Our SACN training process}
 \label{algorithm:SelfAttentionCapsNetAlgo}
\end{algorithm}

\section {Experiments}
\subsection {Medical datasets}

We conducted extensive experiments on highly diverse medical data, and present initial results for natural data as well (as described in the "Natural datasets" subsection). The medical dataset is composed of three separate subsets of images, each contains cancer lesions that are located at different body organs and were screened by different imaging modalities. Two subsets were collected by radiologists at Stanford hospital (250 CT Lung images and 369 MR Brain tumors) and the third set of 1102 CT Liver lesions, is a public one (LiTS). In addition to the differences in the organs and the imaging modalities, these datasets are different from each other also regarding other acquisition criteria; 1) their spatial resolution is within the range of $0.78 mm/pixel-0.94 mm/pixel$ and 2) their slice thickness ranges from $2.5 mm$ to $5 mm$. These differences affect the appearance of the cancer lesions, characterized by a different noise level, homogeneity or contrast relative to the surrounding normal tissue. Each subset has its major challenges but the CT Lung dataset is considered the more difficult dataset for patch classification due to low-contrast lesions and to the similarity to the lung blood vessels, while CT Liver is the easiest one. The inter- and intra-variability between sets of images is shown in Figure \ref{LiTS2}
and in Figure \ref{MedImg}. An external expert annotated two separate regions in each image - normal tissue and cancer lesion. Thirty patches were extracted from each region, means that each training image supplied 60 samples to the whole training cohort. For all experiments, we used $80\%$ of the dataset for training, $10\%$ for testing and $10\%$ percent for validation.
\subsection {Performance evaluation - Medical domain}
The performance of our method was measured as a patch-wise classification - normal or lesion patches. To evaluate the capabilities of our proposed method, we compared the developed architecture with 1) the baseline CapsNet that this work mainly aims to improve, and with 2) the state of the art ResNet-18 and DenseNet-40 architectures. The DenseNet-40 and ResNet-18 were adjusted for being able to analyze small image patches. Similar to what was done when applying ResNet for analysis of CIFAR10 images, we removed some max-pooling layers to ensure that the information will not be lost, prevents the receptive field from shrinking too quickly. We did not want to up-sample the patches too much because it can add noise, which limits the overall performance. Therefore, we carefully considered the up-sampling/max-pooling trade-off and chose the one that supplied the best performacnce. We evaluated the effectiveness of these methods by calculating several statistics. Statistical significance between the methods was calculated by using Wilcoxon paired test.  

\section {Results}
\subsection {Qualitative evaluation}
\begin{figure*}[t!]
  \includegraphics[width=\textwidth]{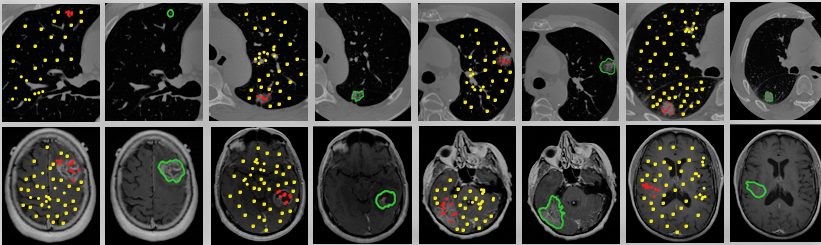}
      \caption{Classification of selected patches. Each pair of images (in the same row) represents the original image with the radiologist's lesion annotation (green) and the processed image. Red - classified lesion patches, Yellow - classified normal patches. Upper row - CT Lung, Bottom row - MR Brain. In each image, we show classification results for the example colored patches.}
      \label{MedImg}
\end{figure*}

Figure  \ref{MedImg} shows the classification results of a subset of randomly chosen patches. For the purpose of visualization, only the colored patches have been classified into normal/lesion regions. The figure shows the substantial diversity of the image characteristics, within and across subsets (CT Lung, CT Liver and MR Brain). Our method shows its ability to handle small lesions, highly heterogeneous lesions and low contrast lesions. It can also distinguish very well between normal structures within the tissue (e.g. blood vessels in CT lung, normal structures in the MR Brain image) and cancer lesions. All these challenges, which usually fail common techniques, are dealt well by our proposed method.
 
\subsection {Quantitative evaluation and Comparison with other common techniques}

Table \ref{table:1} presents the classification accuracy of \textit{all} patches in the testing set (contrary to the subset of patches that is visualized in figure \ref{MedImg}). We set the optimal parameters for every architecture we compared our SACN with, to ensure that any difference between the performances is directly related to the novelty of our architecture. 
Table \ref{table:1} shows that our method  outperforms all other methods for each subset that has been analyzed and for each of the following parameters that has been explored. \textit{First}, the  performance accuracy within each specific subset (Liver, Lung, Brain) is consistently higher when using our proposed method. \textit{Second}, the standard deviation (std) of the classification accuracy over different images within the same subset is lower than the equivalent values when using the baseline CapsNet, DenseNet-40 and ResNet-18. \textit{Third}, the robustness and the stability across different subsets are also significantly higher when using our model.

\begin{table*}[t]
  \centering
  \begin{tabular}{|p{3.05cm}|p{3.05cm}|p{3.05cm}|p{3.05cm}|p{3.05cm}|}
 \hline
 \centering Dataset & \centering ResNet-18 \cite{DBLP:journals/corr/HeZRS15} & \centering DenseNet \cite{DenseNet}  & \centering Baseline CapsNet \cite{sabour2017dynamic} & \centering\textbf{Our SACN}\\ \tabularnewline
 \hline
 \centering Liver (LiTS)  & \centering 0.87$\pm$0.02 & \centering 0.87$\pm$0.01 & \centering 0.89$\pm$0.03 &  \centering\textbf{0.9$\pm$0.01}  \\ \tabularnewline
\hline
 \centering Brain  & \centering 0.91$\pm$0.03 & \centering 0.91$\pm$0.02 & \centering 0.91$\pm$0.02 & \centering\textbf{0.94$\pm$0.01}\\ \tabularnewline
 \hline
\centering Lung  & \centering 0.87$\pm$0.08 & \centering 0.88$\pm$0.07 & \centering 0.85$\pm$0.12 & \centering\textbf{0.92$\pm$0.05} \\ \tabularnewline
 \hline
\end{tabular}
\caption{Comparison (mean, std) of our proposed method with baseline CapsNet, DenseNet and ResNet-18 architectures. The mean and the std values were calculated for different images in the subset. Wilcoxon paired test was calculated ($ p<0.001$ for most comparisons). \textbf{The best results for each subset are bolded}.}
\label{table:1}
\end{table*}

It is worth mentioning that the performance difference between our technique and the other methods we compared with, becomes more significant and going along with the level of the data complexity. This key result enhances the strength of our method. For example the difference between our method and the others is larger for CT Lung and smaller for CT Liver.   

\begin{figure}[!hbt]
 \resizebox{\columnwidth}{!}{%
  \includegraphics{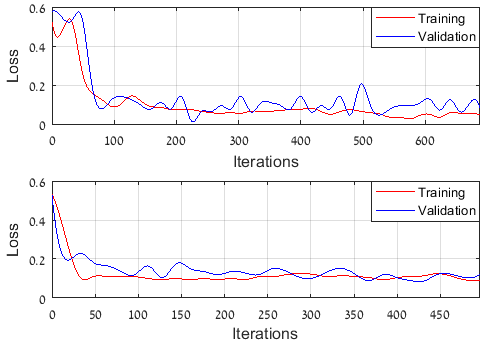}
  }
      \caption{Calculated loss for training and for validation sets. Upper -MR Brain, Bottom - CT Lung.}
      \label{TrainingValidationLoss}
\end{figure}

To ensure that our architecture does not overfit to the training data, we explored the loss/error rates of the training and the validation sets for each individual tested subset (Figure \ref{TrainingValidationLoss}). It can be clearly seen that the loss of the training and the validation sets are comparable, having the same trend and without a substantial differences between them. 

\subsection {Natural datasets}
We showed results for natural data as well, exploring the generalization of the proposed technique to other domains, except for the medical one. The MNIST database includes a training set of 60,000 hand-written digits examples, and a test set of 10,000 examples. The Street View House Numbers (SVHN) is a real-world image dataset. It contains 600,000 digit images that come from a significantly harder real world problem compared to MNIST. The images lack any contrast normalization as well as contain overlapping digits and distracting features which makes them a much more difficult classification dataset compared to MNIST.  CIFAR10 is the third natural dataset that we analyzed by our proposed SACN technique. The dataset consists of 60000 images in 10 classes, with 6000 images per class. There are 50000 training images and 10000 test images.
On contrary to the medical datasets, we used an image-wise classification for natural images. Image-wise analysis was applied because 1) the task was classification of the whole image into a specific class, and because 2) the object and its background in the natural domain were not considered to be too complex ones.\newline We chose a batch size of 64 for MNIST and for CIFAR10 and 32 for SVHN, a learning rate of $2e^{-4}$ and 60 epochs. A weight variance of 0.01 was used. 
Because the natural domain was not the main focus of this work (but it was still important to show generalizability of our model), we compared the performance of our proposed SACN with the baseline CapsNet only, showing the superiority of our methodology on the baseline one. For the MNIST dataset, we obtained a classification accuracy of 0.995, which is comparable to the state of the art methods and to the baseline CapsNet architecture. For the SVHN, we were able to improve the classification accuracy of the baseline CapsNet, which is already pretty high, by $2.4\%$. Lastly, the classification of CIFAR10 was improved by $3.5\%$ by using our SACN, comparing the baseline CapsNet.

\section {Discussion and Conclusion}
\textbf{This paper introduces a novel architecture, called  Self-Attention Capsule Networks (SACN), which was proposed to specifically improve the known CapsNet architecture.} The architecture utilizes the important key ideas of the CapsNet architecture, and boosts its performance by incorporating the Self-Attention mechanism as an integral layer within the CapsNet architecture. Our proposed architecture allows the model parameters, even in shallower layers, to be updated mostly based on image regions that are more relevant to a given task. \newline 
We conducted an extensive set of experiments, focusing on the medical domain but presenting also an analysis of natural images. For the medical subsets, which were part of highly diverse cohort, our proposed method significantly outperformed the baseline CapsNet. We also compared our technique with the advanced state of the art architectures - DenseNet-40 and ResNet-18. Our method was significantly better from these architectures as well. The better performance of our model is reflected in higher accuracy and lower standard deviation. 
Table 1 shows a key advantage of the proposed SACN over the baseline CapsNet, ResNet-18 and DenseNet architectures - \textbf{when the cohort is more complex, the strength of the proposed method becomes more dominant}. This observation is well fitted to the known CapsNet limitation, which tries to account for everything present in an image, and for more complex images, where the background is too diverse, the CapsNet does not perform well in either case. 

In regards to the public natural data that we analyzed, we were able to show classification accuracy that was comparable or better than the CapsNet or other state of the art methods that were reported in literature. 
\textbf{Implementing the model across substantially diverse datasets and domains shows its high generalization, robustness and classification capabilities.} 

The baseline CapsNet is considered an expensive architecture in terms of computational load. For example, analyzing some of the datasets with the baseline CapsNet, resulted in Out of Memory errors on the GPU resources. In this work, we were able to supply classification accuracy that is significantly better than the baseline CapsNet architecture by using a relatively shallow CapsNet architecture and incorporating the attention module. \textbf{We were able to supply better results with less computational load that was reported in literature as a CapsNet cause for process shutdown}. Our architecture is powerful and has potential to be widely-used as it requires less computational resources.

Future work will include additional experiments, focusing on more complex natural and medical datasets. These experiments will be conducted for 2D and 3D data, using additional computational resources.

{\small
\bibliographystyle{ieee}
\bibliography{egbib}
}

\end{document}